\documentclass[a4paper]{article}

\usepackage{INTERSPEECH2020}
\usepackage{multirow}
\usepackage{multicol}
\usepackage{makecell}
\usepackage{amsmath,amssymb}

\usepackage{hyperref}
\hypersetup{
    colorlinks=true,
    linkcolor=black, 
    filecolor=black, 
    urlcolor=blue,
    citecolor=black, 
}

\title{Jointly Encoding Word Confusion Network and Dialogue Context with BERT for Spoken Language Understanding}
\name{Chen Liu$^\star$\thanks{$^\star$ Chen Liu and Su Zhu contributed equally to this work.}, Su Zhu$^\star$, Zijian Zhao, Ruisheng Cao, Lu Chen$^\dagger$ and Kai Yu$^\dagger$\thanks{$^\dagger$ Lu Chen and Kai Yu are the corresponding authors.} \thanks{We thank the anonymous reviewers for their thoughtful comments. This work has been supported by the National Key Research and Development Program of China (Grant No. 2017YFB1002102) and Shanghai Jiao Tong University Scientific and Technological Innovation Funds (YG2020YQ01).}}
\address{
    MoE Key Lab of Artificial Intelligence, AI Institute, Shanghai Jiao Tong University \\
    SpeechLab, Department of Computer Science and Engineering \\
    Shanghai Jiao Tong University, Shanghai, China}

\email{\{chris-chen,paul2204,1248uu,211314,chenlusz,kai.yu\}@sjtu.edu.cn}

\begin{document}

\maketitle
\begin{abstract}
Spoken Language Understanding (SLU) converts hypotheses from automatic speech recognizer (ASR) into structured semantic representations. ASR recognition errors can severely degenerate the performance of the subsequent SLU module. To address this issue, word confusion networks (WCNs) have been used as the input for SLU, which contain richer information than 1-best or n-best hypotheses list. To further eliminate ambiguity, the last system act of dialogue context is also utilized as additional input. In this paper, a novel BERT based SLU model (WCN-BERT SLU) is proposed to encode WCNs and the dialogue context jointly. It can integrate both structural information and ASR posterior probabilities of WCNs in the BERT architecture. Experiments on DSTC2, a benchmark of SLU, show that the proposed method is effective and can outperform previous state-of-the-art models significantly.

\end{abstract}

\section{Introduction}
\label{sec:intro}

The spoken language understanding (SLU) module is a key component of spoken dialogue system (SDS), parsing user utterances into corresponding semantic representations (e.g., dialogue acts \cite{young2007cued}). For example, the utterance ``\emph{I want a high priced restaurant which serves Chinese food}'' can be parsed into a set of semantic tuples ``\emph{inform(price$\_$range=expensive), inform(food=Chinese)}''. In this paper, we focus on SLU with semantic labels in the form of \emph{act(slot=value)} triplets (i.e., unaligned annotations), which does not require word by word annotations. Both discriminative \cite{Mairesse2009SpokenLU,henderson2012discriminative,zhu2014semantic,barahona2016exploiting} and generative \cite{zhao2018improving,zhao2019hierarchical} methods have been developed to extract semantics from ASR hypotheses of the user utterance. 

SLU systems trained on manual transcripts would get a dramatic decrease in performance when applied to ASR hypotheses \cite{mesnil2013investigation}. To eliminate ambiguity caused by ASR errors, two kinds of input features can be exploited to enhance SLU models: (1) ASR hypotheses and (2) dialogue context information. 1) Considering the uncertainty of ASR hypotheses, previous works utilized ASR 1-best result \cite{henderson2012discriminative,zhao2019hierarchical,zhu2018robust,li2019robust}, N-best lists \cite{henderson2012discriminative,robichaud2014hypotheses, khan2015hypotheses}, word lattices \cite{vsvec2015word,ladhak2016latticernn,huang2019adapting} and word confusion networks (WCNs) \cite{hakkani2006beyond,henderson2012discriminative,tur2013semantic,yang2015using,jagfeld2017encoding,masumura2018neural} for inputs to train an SLU model. Masumura et al. \cite{masumura2018neural} proposed a fully neural network based method, \emph{neural ConfNet classification}, to encode WCNs. It first obtains bin (each bin contains multiple word candidates and their posterior probabilities of ASR hypothesis in the same time step) vectors by the weighted sum of all word embeddings in each bin separately, and then exploits a bidirectional long short-term memory recurrent neural network (BLSTM-RNN) to integrate all bin vectors into an utterance vector. Nevertheless, bin vectors are extracted locally, ignoring contextual features beyond certain bins. 2) Furthermore, the last system dialogue act~\cite{young2007cued} can be utilized to track context~\cite{henderson2012discriminative,barahona2016exploiting}, and provide some implications about the user intent under noisy conditions. However, utterance and context are independently encoded by different models to generate the final representation, resulting in a lack of interaction between them.



Recently, pre-trained language models, such as GPT \cite{radford2018improving} and BERT \cite{devlin2018bert}, have been successfully adopted in various NLP tasks. Huang et al. \cite{huang2019adapting} adapted GPT for modeling word lattices, where lattices are represented as directed acyclic graphs. However, GPT is modeled as a unidirectional Transformer~\cite{vaswani2017attention} and neglects context in the future, thus less expressive than BERT. Although both word lattices and WCNs contain more information than N-best lists, WCNs have been proven more efficient in terms of size and structure~\cite{hakkani2006beyond}. 

To these ends, we propose a novel BERT (Bidirectional Encoder Representations from Transformers)~\cite{devlin2018bert} based SLU model to jointly encode WCNs and system acts, which is named WCN-BERT SLU. It consists of three parts: a BERT encoder for jointly encoding, an utterance representation model, and an output layer for predicting semantic tuples. The BERT encoder exploits posterior probabilities of word candidates in WCNs to inject ASR confidences. Multi-head self-attention is applied over both WCNs and system acts to learn context-aware hidden states. The utterance representation model produces an utterance-level vector by aggregating final hidden vectors. Finally, we add both discriminative and generative output layers to predict semantic tuples. To the best of our knowledge, this is the first work to leverage the structure and probabilities of input tokens in BERT. Our method is evaluated on DSTC2 dataset \cite{henderson2014second}, and the experimental results show that our method can outperform previous state-of-the-art models significantly. 


\section{WCN-BERT SLU}
\label{sec:model}

\begin{figure*}[t]
    \centering
    \includegraphics[width=0.88\textwidth]{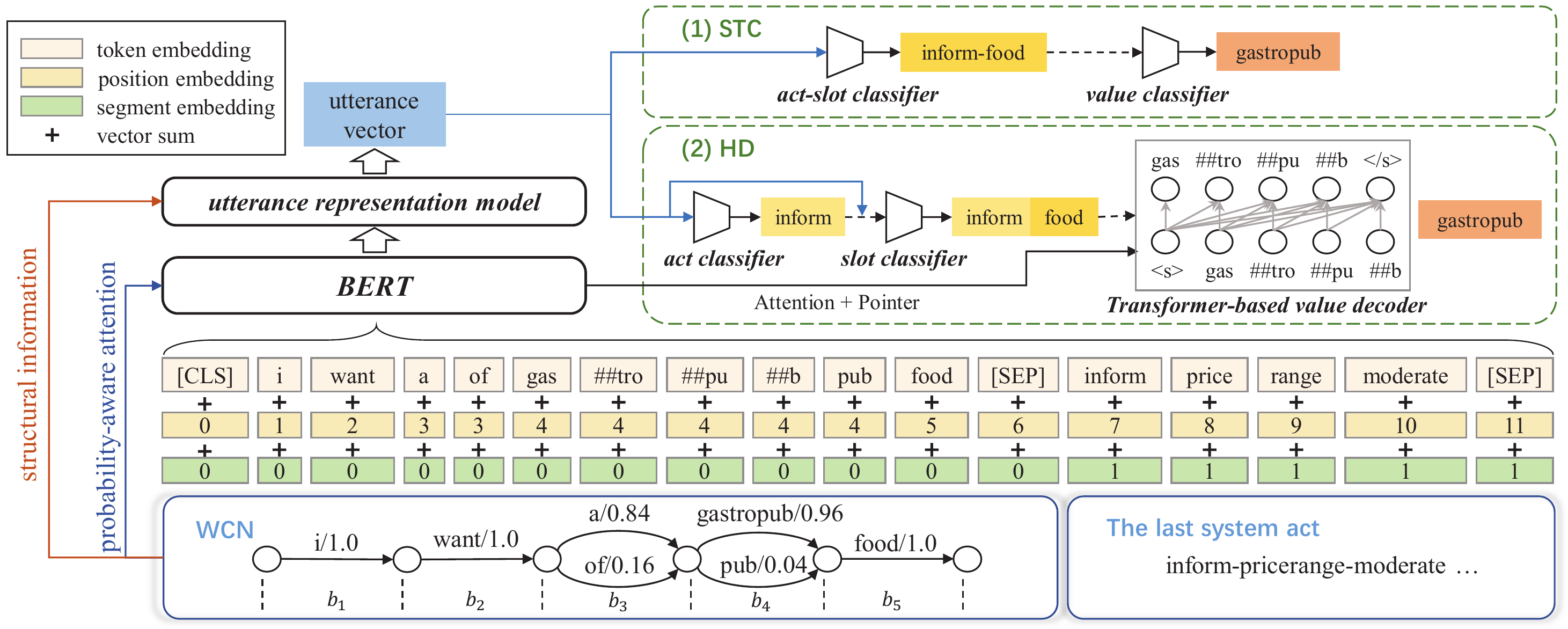}
    \caption{An overview of our proposed WCN-BERT SLU, which contains a BERT encoder, an utterance representation model, and an output layer. First, the WCN and the last system act are arranged as a sequence to be fed into the BERT encoder. The token-level BERT outputs are then integrated into an utterance-level vector representation. Finally, either a discriminative (semantic tuple classifier, STC) or generative (hierarchical decoder, HD) approach is utilized in the output layer for predicting the act-slot-value triplets.}
    \label{fig:model_arch_full}
    \vspace{-0.5em}
\end{figure*}

In this section, we will describe the details of the proposed framework, as shown in Fig. \ref{fig:model_arch_full}. For each user turn, the model takes the corresponding WCN and the last system act as input, and predicts the semantic tuples at turn level. 

\subsection{Input representation}
\label{subsec:input_repr}

The WCN is a compact lattice structure where candidate words paired with their associated posterior probabilities are aligned at each position \cite{feng2009effects}. It is commonly considered as a sequence of word bins $B=(b_1,\cdots,b_{M})$. The $m$-th bin can be formalized as $b_m=\{(w_m^1, P(w_m^1)),\cdots,(w_m^{I_m},P(w_m^{I_m}))\}$, where $I_m$ denotes the number of candidates in $b_m$, $w_m^i$ and $P(w_m^i)$ are respectively the $i$-th candidate and its posterior probability given by ASR system. The WCN is flattened into a word sequence, $w^{\text{WCN}}=(w_1^1,\cdots,w_1^{I_1},\cdots,w_{M}^1,\cdots, w_{M}^{I_{M}})$. 

System acts contain dialogue context in the form of ``act-slot-value'' triplets. We consider the last system act just before the current turn, $A=((a_1,s_1,v_1), \cdots, (a_K,s_K,v_K))$, where $K$ is the number of triplets, $a_k$, $s_k$ and $v_k$ are the $k$-th act, slot and value, respectively. It is also arranged as a sequence $w^{\text{SA}}=(a_1, s_1, v_1, \cdots, a_K, s_K, v_K)$. Act and slot names are tokenized, e.g., ``pricerange'' is split into ``price'' and ``range''.

To feed WCNs and system acts into BERT together, we denote the input token sequence as $w=(w_1,\cdots,w_T)=\texttt{[CLS]} \oplus \textsc{Tok}(w^{\text{WCN}}) \oplus \texttt{[SEP]} \oplus \textsc{Tok}(w^{\text{SA}}) \oplus  \texttt{[SEP]}$, where $\oplus$ concatenates sequences together, $\textsc{Tok}(\cdot)$ is the BERT tokenizer which tokenizes words into sub-words, $\texttt{[CLS]}$ and $\texttt{[SEP]}$ are auxiliary tokens for separation, $T = |\textsc{Tok}(w^{WCN})| + |\textsc{Tok}(w^{SA})| + 3$. Considering the structural characteristics of WCN, i.e., multiple words compete in a bin, we define that all words (in fact sub-words after tokenization) in the same bin share the same position ID.

Eventually, the BERT input layer embeds $w$ into $d_x$-dimensional continuous representations by summarizing three embeddings \cite{devlin2018bert} as follows:
\begin{align}
{x}_t = E_{\text{token}}(w_t) + E_{\text{pos}}(w_t) + E_{\text{seg}}(w_t), {x}_t \in \mathbb{R}^{d_x}
\end{align}
where $E_{\text{token}}(\cdot)$, $E_{\text{pos}}(\cdot)$ and $E_{\text{seg}}(\cdot)$ stand for WordPiece \cite{wu2016google}, positional and segment embeddings, respectively.

\subsection{BERT encoder}

BERT consists a series of bidirectional Transformer encoder layers \cite{vaswani2017attention}, each of which contains a multi-head self-attention module and a feed-forward network with residual connections \cite{he2016deep}. For the $l$-th layer, assuming the input is $x^{l}=(x_1^{l}, \cdots, x_T^{l})$, the output is computed via the self-attention layer (consisting of $H$ heads):
\allowdisplaybreaks
\begin{align*}
e_{i j}^{l,h} &=\frac{(W_{Q}^{l,h}{x}_{i}^l)^{\top}(W_{K}^{l,h} {x}_{j}^l)}{\sqrt{d_{x} / H}} ; \quad \alpha_{i j}^{l,h}=\frac{\exp (e_{i j}^{l,h})}{\sum_{k=1}^{T} \exp (e_{i k}^{l,h})}\\
{z}_{i}^{l,h} &=\sum_{j=1}^{T} \alpha_{i j}^{l,h}( W_{V}^{l,h} {x}_{j}^l); \ \  {z}_{i}^l=\operatorname{Concat}({z}_{i}^{l,1}, \cdots, {z}_{i}^{l,H}) \\
\tilde{{x}}_{i}^{l+1} &=\operatorname{LayerNorm}({x}_{i}^l+\operatorname{FC}({z}_{i}^l)) \\
{x}_{i}^{l+1} &=\operatorname{LayerNorm}(\tilde{{x}}_{i}^{l+1}+\operatorname{FC}(\operatorname{ReLU}(\operatorname{FC}(\tilde{{x}}_{i}^{l+1}))))
\end{align*}
where FC is a fully-connected layer, LayerNorm denotes layer normalization~\cite{ba2016layer}, $1\leq h \leq H$, $W_Q^{l, h},W_K^{l, h},W_V^{l, h}\in\mathbb{R}^{(d_x/H)\times d_x}$.

\noindent \textbf{WCN probability-aware self-attention} \quad We propose an extension to the self-attention mechanism to consider the posterior probabilities of tokens in WCN. Following the notations in section \ref{subsec:input_repr}, for each input token sequence $w=(w_1, \cdots, w_T)$, we define its corresponding probability sequence as $p=(p_1, \cdots, p_T)$, with
\begin{align}
p_t = \begin{cases}
P(w_t) & w_t \in \textsc{Tok}(w^{\text{WCN}}) \\
1.0 & \text{otherwise}
\end{cases}
\end{align}
where $P(w_t)$ denotes the ASR posterior probability of a token. Note that the probability of a sub-word equals that of the original word in WCN. Probabilities of tokens in the system acts are defined as $1.0$. Now we inject ASR posterior probabilities into the BERT encoder by changing the computation of $e_{ij}^{l, h}$:
\begin{align}
e_{ij}^{l, h} &= (W_Q^{l, h}x_i^{l})^{\top}(W_K^{l, h}x_j^{l})/\sqrt{d_x/H} + \lambda^{l, h}\cdot p_j
\end{align}
where $\lambda^{l, h}$ is a trainable parameter. 

Finally, token-level representations are produced after the stacked encoder layers, denoted as $o=(o_1,\cdots,o_{T})$.

\begin{figure}[htbp]
    \centering
    \includegraphics[width=0.8\linewidth]{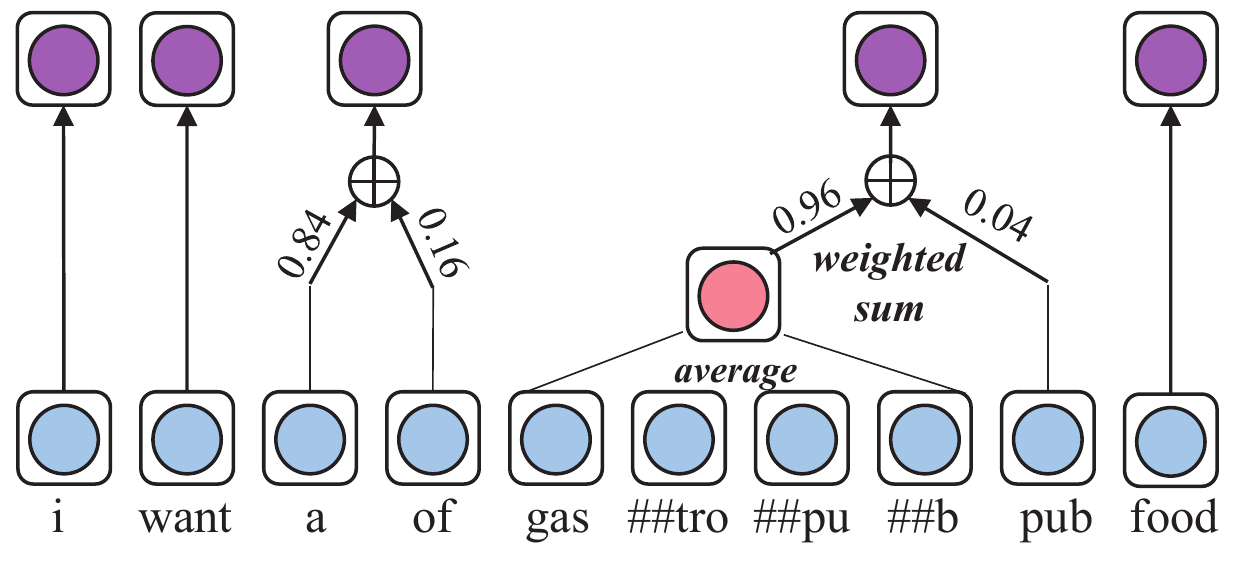}
    \caption{Illustration of the feature aggregation. Only the WCN part of an input token sequence is presented. The token-level features are in blue while bin-level features are in purple.}
    \label{fig:utt_repr}
\end{figure}

\subsection{Utterance representation}

The output hidden vector corresponding to the \texttt{[CLS]} token, i.e., $o_1$, is usually applied to represent the whole utterance. Additionally, we propose to gather the other hidden vectors by considering the structural information of the WCN.


Firstly, the token-level hidden vectors of the WCN part are aggregated into bin-level through the following two steps: (1) BERT sub-word vectors belong to an input word are averaged to be the word vector; (2) word-level vectors of each bin are then weighted and summed to get a bin vector. An example of the feature aggregation of the WCN part is illustrated in Fig. \ref{fig:utt_repr}, while the features corresponding to the system acts are unchanged. After aggregation, we get new feature vectors $u=(u_1, \cdots, u_{T'})$, where $T' = M + |\textsc{Tok}(w^{SA})| + 3$ ($M$ is the number of bins) and $T' \leq T$.


Then, we summarize the bin-level features with a self-attentive approach as follows:
\begin{align}
u'=\sum_{t=2}^{T'}{\left[\dfrac{\exp(\Bar{v}^{\top}\tanh(W_a u_t+b_a))}{\sum_{j=2}^{T'}\exp(\Bar{v}^{\top}\tanh(W_a u_j+b_a))} \cdot u_t\right]} \label{eqn:self_attn}
\end{align}
where $b_a, \Bar{v} \in \mathbb{R}^{d_x}$ and $W_a \in \mathbb{R}^{d_x\times d_x}$ are trainable parameters. The final utterance representation is obtained by concatenating $u'$ with the hidden state of \texttt{[CLS]}, i.e., $r=\operatorname{Concat}(u_1, u')$.

\subsection{Output layer}

To evaluate the validity and portability of the encoding model, we apply both discriminative (Sec. \ref{sec:stc}) and generative (Sec. \ref{sec:hd}) approaches for predicting the act-slot-value labels.

\subsubsection{Semantic tuple classifier (STC)}
\label{sec:stc}

Upon the final utterance representation $r$, we apply a binary classifier for predicting the existence of each act-slot pair, and a multi-class classifier over all possible values existing in the training set for each act-slot pair~\footnote{For act-slot pairs requiring no value (like \emph{thankyou, goodbye, request-phone, request-food} \emph{etc.}), the value classification is omitted.}. Therefore, this method cannot predict values unseen in the training set.



\subsubsection{Transformer-based hierarchical decoder (HD)}
\label{sec:hd}

To improve the generalization capability of value prediction, we follow Zhao et al.~\cite{zhao2019hierarchical} to construct a hierarchical decoder consisting of an act classifier, a slot classifier, and a value generator. However, there are two main differences listed as follows:

1) Acts and slots are tokenized and then embedded by the BERT embedding layer as additional features for the slot classifier and the value generator. For each act $a$ or slot $s$, sub-word level vectors from BERT embedding layer are averaged to get a single feature vector ($e_a$ for $a$ and $e_s$ for $s$).

2) We replace the LSTM-based value generator with a Transformer-based one~\cite{vaswani2017attention}. Tokenized values are embedded by the BERT embedding layer and generated at the sub-word level. Therefore, we tie the BERT's token embeddings with the weight matrix of the linear output layer in the value generator. 


\section{Experiments}
\label{sec:exp}

We experiment on the dataset from the second Dialog State Tracking Challenge (DSTC2) \cite{henderson2014second}, which contains $11677$, $3934$, $9890$ utterances for training, validation and testing, respectively. In order to shorten flattened WCN sequences, we prune WCNs by removing interjections~\footnote{Such as \emph{uh}, \emph{oh}, etc. according to Jagfeld et al. \cite{jagfeld2017encoding}.} and word candidates with probabilities below a certain threshold ($0.001$ as recommended \cite{jagfeld2017encoding}). The evaluation metrics are F$_1$ score of \emph{act-slot-value} triplets and utterance-level accuracy. We do not assume that all value candidates of each slot are known in advance. 

In our experiments, we use English uncased BERT-base model, which has $12$ layers of $768$ hidden units, and 12 attention heads. During training, Adam \cite{kingma2014adam} is used for optimization. We select the initial learning rate from \{5e-5, 3e-5, 2e-5\}, with a warm-up rate of $0.1$ and an L2 weight decay of $0.01$. The maximum norm for gradient clipping is set to $1$. The dropout rate is set to $0.1$ for the BERT encoder and $0.3$ for the utterance representation model and the output layer. The model is trained for $50$ epochs and saved according to the best performance on the validation set. Each experimental setting was run five times with different random seeds, and we report the average result. \footnote{The code is available at \href{https://github.com/simplc/WCN-BERT}{github.com/simplc/WCN-BERT}.}

\subsection{Main results}

\begin{table}[htbp]
    \caption{F$_1$ scores (\%) and utterance-level accuracies (\%) of baseline models and our proposed model on the test set.}
    \label{tab:main_res}
    \small
    \centering
    \begin{tabular}{l|p{0.5cm}|l||cc}
        \hline
        \multicolumn{2}{l|}{\textbf{Models}} & \textbf{Train}$\to$\textbf{Test}  & \textbf{F$_1$} & \textbf{Acc.} \\
        \hline\hline
        \multirow{5}{*}{\makecell[l]{BLSTM+\\Self-Attn}} & \multirow{5}{*}{STC} & manual$\to$manual & 98.56 & 97.29 \\
        & & manual$\to$1-best & 83.57 & 74.91 \\
        &  & 1-best$\to$1-best & 84.06 & 75.26 \\
        &  & 1-best$\to$10-best & 84.92 & 75.70 \\
        &  & 10-best$\to$10-best & 85.05 & 77.07 \\
        \hline
        \makecell[l]{Neural\\ConfNet} & STC & WCN$\to$WCN & 85.01 & 77.03 \\
        \hline
        Lattice-SLU & STC & WCN$\to$WCN & 86.09 & 78.78 \\
        \hline
         \multirow{2}{*}{\makecell[l]{WCN-BERT\\(ours)}} & STC & \multirow{2}{*}{WCN$\to$WCN}  & \textbf{87.91}  & \textbf{81.14} \\
         & HD & & 87.33 & 80.74 \\
        \hline
    \end{tabular}
    \vspace{-0.5em}
\end{table}

As shown in Table \ref{tab:main_res}, different types of inputs are applied for baselines, including \emph{manual} transcriptions, ASR \emph{1-best}, \emph{10-best} lists, and \emph{WCNs}. The baseline model for the first three types (manual, 1-best, and 10-best) is BLSTM with self-attention~\cite{masumura2018neural}.~\footnote{In these baselines, word embedding is initialized with 100-dim Glove6B \cite{pennington2014glove}. The learning rate is set to $0.001$, fixed during training. The maximum norm for gradient clipping is $5$, and dropout rate is $0.3$.} BLSTM encodes the input sequence and gets the utterance representation $r$ with the self-attention similar to Eq. (\ref{eqn:self_attn}). To test on 10-best lists with the model trained on 1-best, we run the model on each hypothesis from the list and average the results weighted by the ASR posterior probabilities. For direct training and evaluation on 10-bests, the representation vector $r$ is calculated as $r=\sum_{i=1}^{10}{\gamma_i r_i}$, where $r_i$ is the representation vector of hypothesis $i$ and $\gamma_i$ is the corresponding ASR posterior probability.

For modeling WCNs, we follow the \emph{Neural ConfNet Classification} \cite{masumura2018neural} method. WCNs are fed into the model through a simple weighted sum representation method, where all word vectors are weighted by their posterior probabilities and then summed. We also apply the \emph{Lattice-SLU} method with GPT \cite{huang2019adapting} on WCNs. The output layer of all baselines is STC (Sec. \ref{sec:stc}).

By comparing across the baselines, we find that performances become better with larger ASR hypotheses space. The \emph{Neural ConfNet Classification} method can outperform the system trained and tested with 1-best, and it achieves comparable results to the 10-best system. With powerful pre-trained language models (GPT~\cite{radford2018improving}), the \emph{Lattice-SLU} beats the other baselines above. The last system act is not exploited in the baselines, which will be analyzed in the following ablation study.

\begin{table}[htbp]
    \caption{Comparison with prior arts on the DSTC2 dataset.}
    \label{tab:art_res}
    \small
    \centering
    \begin{tabular}{l||c}
        \hline
        \textbf{Models} & \textbf{F$_1$ (\%)} \\
        \hline\hline
        \textsc{SLU2} \cite{williams2014web} & 82.1  \\
        \textsc{CNN+LSTM\_w4} \cite{barahona2016exploiting} & 83.6  \\
        \textsc{CNN} \cite{zhao2018improving} & 85.3 \\
        \textsc{S2S-Attn-Ptr-Net} \cite{zhao2018improving} & 85.8  \\
        \textsc{Hierarchical Decoding} \cite{zhao2019hierarchical} & 86.9  \\
        \hline
        WCN-BERT + STC (ours) & \textbf{87.9} \\
        \hline
    \end{tabular}
    \vspace{-0.5em}
\end{table}

By joint modeling WCNs and the last system act with powerful pre-trained BERT, our proposed framework outperforms the baselines significantly in F$_1$ score and utterance-level accuracy, and achieves new state-of-the-art performance on the DSTC2 dataset, as shown in Table \ref{tab:main_res} and Table \ref{tab:art_res}.

\subsection{Ablation study}

In this section, we perform ablation experiments of our WCN-BERT SLU with both STC and HD, as presented in Table \ref{tab:ablation}.



\begin{table}[htbp]
    \caption{F$_1$ scores (\%) of ablation study on the test set.}
    \label{tab:ablation}
    \small
    \centering
    \begin{tabular}{ll||cc}
        \hline
        &\textbf{Model variations} & \textbf{STC} & \textbf{HD} \\
        \hline\hline
        (a) & WCN-BERT & 87.91 & 87.33   \\
        \hline
        (b) & \quad w/o WCN Prob & 87.15 & 86.47  \\
        (c) & \quad $r=u_1$ & 87.75 & 86.99 \\
        \hline
        (d) & \quad w/o BERT & 86.12 & 86.01 \\
        (e) & \quad w/o system act & 86.69 & 86.18 \\
        (f) & \quad w/o BERT and system act & 85.85 & 85.06 \\
        \hline 
        (g) & \quad \makecell[l]{Replace system act with\\ system utterance} & 88.01 & 87.28 \\
        \hline
    \end{tabular}
    \vspace{-0.5em}
\end{table}

Row (b) shows the results without considering WCN probabilities in the BERT encoder. In this case, the model lacks prior knowledge of ASR confidences, resulting in a significant performance drop ($0.76\%$ for STC and $0.86\%$ for HD). By only considering the hidden state related to \texttt{[CLS]} as the utterance representation (row(c)), i.e., $r=u_1$, the F$_1$ scores decrease with both STC and HD, indicating that the structural information of WCN is beneficial for utterance representation.

By removing BERT (row (d)), we utilize vanilla Transformer to jointly encode WCNs and system acts but not fine-tune a pre-trained BERT. We use 100-dim word embeddings, initialized with Glove6B \cite{pennington2014glove}. The results show that removing BERT brings about a remarkable decrease in F$_1$ score ($1.79\%$ for STC and $1.32\%$ for HD).

Besides, we investigate the effect of dialogue context by removing the last system act from the input, as demonstrated in row (e). Results show that jointly encoding the WCN and the last system act improves the performance dramatically. It is a fair comparison with the \emph{Lattice-SLU} baseline in Table~\ref{tab:main_res}, both with pre-trained language models. The result implies that our model is much more effective, owing to (1) the capability of the bidirectional Transformer, which considers the future context, and (2) the different considerations of WCN structures.

In row (f), neither BERT nor the dialogue context is included. This also gives a fair comparison with the \emph{Neural ConfNet Classification} method \cite{masumura2018neural} at the model level (Transformer v.s. BLSTM). With STC as the output layer, the Transformer with WCN probability-aware self-attention mechanism is shown to have better modeling capability (an improvement of 0.84\% in F$_1$) and higher training efficiency (over $6\times$ faster).

Moreover, we replace the last system act with the last system utterance in the input (row (g)), which only causes a slight change (within $0.1\%$) in F$_1$ scores. This indicates that our model is also applicable to datasets with only system utterances, but no system acts.

\subsection{Analysis of hierarchical decoder}

As we can see from the previous results, models with hierarchical decoder (HD) perform worse than STC. However, the generative approach equips the model with generalization capability, and the pointer-generator network is beneficial to handle out-of-vocabulary (OOV) tokens. To analysis the generalization capability of the proposed model with hierarchical decoders, we randomly select a certain proportion of the training set to train our proposed WCN-BERT SLU with STC or HD. The validation and test set remain unchanged. Furthermore, we evaluate the F$_1$ scores of \texttt{seen} and \texttt{unseen} act-slot-value triplets, according to whether a triplet is seen in the training set.


\begin{table}[htbp]
    \caption{F$_1$ scores (\%) of our WCN-BERT SLU on the test set with varying training size.}
    \label{tab:hd_analysis}
    \small
    \centering
    \begin{tabular}{c|l||ccc}
        \hline
        \textbf{\makecell[c]{Train size}} & \textbf{Models} & overall & \texttt{seen} & \texttt{unseen} \\
        \hline\hline
         \multirow{2}{*}{1\%} &  + STC & 62.9 & 67.1 & 0.0 \\
         &   + HD & \textbf{71.6} & \textbf{77.0} & \textbf{29.7} \\
        \hline
         \multirow{2}{*}{5\%} &   + STC & 78.7 & 80.2 & 0.0 \\
         &   + HD & \textbf{81.5} & \textbf{83.1} & \textbf{35.3} \\
        \hline
         \multirow{2}{*}{10\%} &   + STC & 81.3 & 81.9 & 0.0 \\
         &   + HD & \textbf{82.9} & \textbf{83.8} & \textbf{22.9} \\
        \hline
         \multirow{2}{*}{100\%} &   + STC & \textbf{87.9} & \textbf{88.2} & 0.0 \\
         &   + HD & 87.3 & 87.5 & \textbf{4.8} \\
        \hline
    \end{tabular}
    \vspace{-0.5em}
\end{table}

As shown in Table \ref{tab:hd_analysis}, with the training size getting decreased, the overall performance of HD will not degrade sharply. Moreover, the hierarchical decoder is shown to have better generalization capability in the face of unseen labels.
\section{Conclusion}
\label{sec:conclusion}

In this paper, we propose to jointly encode WCN and dialogue context with BERT for SLU. To eliminate ambiguity caused by ASR errors, WCNs are utilized for involving ASR hypotheses uncertainties, and dialogue context implied by the last system act is exploited as auxiliary features. In addition, the pre-trained language model BERT is introduced to better encode WCNs and system acts with self-attention. Experimental results show that our method can beat all baselines and achieves new state-of-the-art performance on DSTC2 dataset. Except for encoding different ASR hypotheses, there are several studies about ASR error correction for SLU~\cite{schumann2018incorporating,weng2020joint}, which will be our future work.

\bibliographystyle{IEEEtran}

\bibliography{mybib}

\end{document}